\documentclass[a4paper]{llncs}
\usepackage[T1]{fontenc}
\usepackage{ifpdf}
\usepackage{cite}
\usepackage{hyperref}
\ifpdf
  \usepackage[pdftex]{graphicx}
  \graphicspath{{images/pdf/}{images/jpeg/}{images/png/}}
  \DeclareGraphicsExtensions{.pdf,.jpeg,.png}
\else
  \usepackage[dvips]{graphicx}
  \graphicspath{{images/eps/}{images/png/}}
  \DeclareGraphicsExtensions{.eps}
\fi
\usepackage{subfig}
\usepackage{dsfont}

\begin{document}
\title{Smooth Symbolic Regression:\\Transformation of Symbolic Regression into a Real-valued Optimization Problem}
\titlerunning{Smooth Symbolic Regression}
\toctitle{Smooth Symbolic Regression: Transformation of Symbolic Regression into a Real-valued Optimization Problem}\footnotetext[1]{The final publication is available at \url{https://link.springer.com/chapter/10.1007/978-3-319-27340-2_47}}
\author{Erik Pitzer \and Gabriel Kronberger}
\authorrunning{E. Pitzer and G. Kronberger}
\tocauthor{Erik Pitzer, and Gabriel Kronberger}
\institute{Heuristic and Evolutionary Algorithms Laboratory,
School of Informatics, Communications and Media,
University of Applied Sciences Upper Austria,
Franz-Fritsch Strasse 11, 4600 Wels, AUSTRIA,
\email{\{erik.pitzer,gabriel.kronberger\}@fh-hagenberg.at},
\texttt{http://heal.heuristiclab.com} and \texttt{http://fh-ooe.at}}

\maketitle

\begin{abstract}
  The typical methods for symbolic regression produce rather abrupt changes in
  solution candidates. In this work, we have tried to transform symbolic
  regression from an optimization problem, with a landscape that is so rugged
  that typical analysis methods do not produce meaningful results, to one that
  can be compared to typical and very smooth real-valued problems. While the
  ruggedness might not interfere with the performance of optimization, it
  restricts the possibilities of analysis. Here, we have explored different
  aspects of a transformation and propose a simple procedure to create
  real-valued optimization problems from symbolic regression problems.
 \end{abstract}

\section{Introduction}
When analyzing complex data sets not only to predict a variable from others
but also trying to gain insights into the relationships between variables,
symbolic regression is a very valuable tool. It can often produce
human-interpretable explanations for relationships between variables. The
considered formulas are explored by substituting variables or re-organizing
the syntax tree. This results in rather abrupt changes in the behavior of
these formulas.

Genetic programming (GP) \cite{Koza1992} is a technique that uses an 
evolutionary algorithm to evolve computer programs that solve a given problem 
when executed.  These programs are often represented as symbolic expression 
trees, and operations such as crossover and mutation are performed on 
sub-trees.  One particular problem that can be solved by GP is symbolic 
regression \cite{Koza1992}, where the goal is to find a function, mapping the 
known values of input variables to the value of a target variable with minimal 
error.

Several specialized and improved variants of GP for symbolic regression have 
already been described in the literature
e.g. \cite{Keijzer2003,Kotanchek2007,Vladislavleva2009,Kommenda2013} 
and it has been shown that other techniques are also viable for solving 
symbolic regression problems such as FFX \cite{McConaghy2011} or
dynamic programming \cite{Worm2013}.


While GP -- or other (quasi-)combinatorial methods -- typically work well in
identifying formulas that describe relations between variables and, therefore,
do not create a black-box function but an opaque and often human-interpretable
description of these relations, the process for obtaining these formulas is
complicated and and large effort is necessary to grasp their progression.

During the optimization, the symbolic expression tree can vary wildly between 
individuals and between generations. In other words, when a GP fitness 
landscape is analyzed it appears extremely rugged.  A typical measure for 
ruggedness, the autocorrelation \cite{Weinberger1990} between two 
``neighboring'' solution candidates, is usually very close to zero.

This makes it very hard to derive any meaningful conclusions of GP fitness  
landscapes as typical ``neighbors'' are too dissimilar to each other, even more
so, when crossover operators are employed which creates even more drastic
changes, let alone the difficulties of crossover landscapes themselves
\cite{Stadler1998}. 

In this paper, we present the idea of creating a smoother fitness landscape for 
symbolic regression problems by borrowing ideas from neural networks and 
combining them with typical tree formulations found in genetic programming. The
intent is to have a smooth transition from one syntax tree to another. At the
same time, we want to ensure that the final output settles for and determines
one of the available operators at each node.

\subsection{Symbolic Regression}\label{sec:symbolic-regression}
In a regression problem the task is to find a mapping from a set of input
variables to on or more output variables so that using only the input values,
the output can be accurately predicted. This task can be tackled with two
fundamentally different approaches. On the one hand, so-called black-box models
focus on providing predictions with maximum quality sacrificing
``understanding'' of the model for exampling when employing neural networks
\cite{Rosenblatt1958} and deep learning \cite{Deng2013} or Support Vector
Machines \cite{Cortes1995}. On the other hand, white box models try not only to
give good quality explanations but also try to provide some insight into how the
relationship between the variables. Examples are most prominently linear
regression \cite{Freedman2009} and generalized linear regression
\cite{Nelder1972}.

As an extension to these methods, symbolic regression provides great freedom in
the formulation of a regression formula. By allowing an arbitrary syntax tree as
the formulation for the relationship between the variables. This freedom,
however, comes with the price of a much large solution space and, hence, much
more possibilities for a goo solution. Therefore, powerful methods have to be
used to control the complexity of this approach.

Genetic Programming (GP) \cite{Koza1992} is a method that is able to conquer
these problems and create white-box models with good quality and often
understandable models that can give new insights into the relationships between
the variables in addition to the provided predictions.

In Genetic Programming, syntax trees are usually modified using two different
methods both drawing the solution candidates from a pool called the current
generation. The most important form of modification is achieved via crossover,
where two trees are recombined into a new tree that has some features from both
predecessors. The second form of modification is mutation that randomly
introduces or changes some features of the syntax tree, such as replacing
operators or changing constants.


Overall, genetic programming provides powerful means of evolving symbolic trees
and has proven its effectiveness and efficiency in providing good quality white
box models of difficult regression problems. On the downside, however, the
changes introduced during the evolution of syntax trees are quite drastic and
the process, how genetic programming arrives at these solutions is very
difficult to follow through, and its analysis is complicated. 

\subsection{Fitness Landscape Analysis}
Every optimization problem implies a so-called fitness landscape that
describes the relationship of solution candidates and their
associated fitness. Formally, a fitness landscape can be defined as the triple
$\mathcal{F} := \left\{ \mathcal{S}, f, \mathcal{X} \right\}$, where
$\mathcal{S}$ is an arbitrary set of solution candidates for the optimization
problems at hand. The function $f: \mathcal{S} \rightarrow \mathds{R}$ is the
actual fitness function that assigns a value of desirability to every solution
candidate and is often the most expensive part in the optimization of a
problem. Finally, $\mathcal{X}$ describes how solution candidates relate to each
other: A very simple case would be to define $\mathcal{X} \subseteq \mathcal{S}
\times \mathcal{S}$ a relation of neighboring solution candidates or
alternatively as distance function $\mathcal{X} : \mathcal{S} \times \mathcal{S}
\rightarrow \mathds{R}$. The important observation is that this definition of a
fitness landscape can be made for any optimization problem and, therefore, 
provides the foundation for very general and portable problem analysis
techniques.
 
Based on this formulation several different analysis methods have been
proposed. Many of which require a sample of the solution space, i.e. so called
walkable landscapes \cite{Hordijk1996}. This sample often comes in the form of a
trajectory inside the solution space, following the neighborhood relation or
distance function. 

Based on these trajectories, different measures can be defined that characterize
the fitness landscape. Examples are auto correlation \cite{Weinberger1990} that
measures the average decay of correlation of fitness values as the trajectory moves away
from a point. Typically, it is only defined for the very first step but can be
continued to an arbitrary distance. The distance at which the correlation is not
statistically significant anymore is called the correlation length also defined
in \cite{Weinberger1990}.

Other techniques include the information analysis proposed in \cite{Vassilev2000}
where several measure are defined that try to capture information theoretic
characteristics of these trajectories. One particularly simple but interesting
property is the information stability which simply captures the maximum fitness
difference between neighboring solution candiates and has proven to be a very
characteristic property of a problem instance.

Besides these trajectory-based analysis methods several other methods have been
proposed such as analytical decomposition into elementary landscapes
\cite{Stadler1994,Chicano2011} or fitness distance correlation
\cite{Jones1995}.

\section{Transformation}
As described in the previous sections most landscape analysis methods as well as
trajectory-based optimization algorithms rely on a relatively smooth landscape
or, in other words, high correlation between neighboring solution candidates.
Conversely, typical modifications in genetic programming are very large and
previous attempts of applying classic fitness landscape analysis (FLA) have
failed.

\begin{figure}[htbp]
  \centering
  \subfloat[fixed structure]{
    \label{fig:fixed-structure}
    \centering
    \hspace{0.15\textwidth}
    \includegraphics[width=0.20\textwidth]{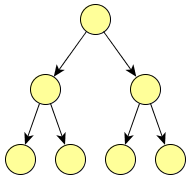}
    \hspace{0.15\textwidth}}
  \subfloat[operation average]{
    \label{fig:smooth-transition}
    \centering
    \hspace{0.15\textwidth}
    \includegraphics[width=0.20\textwidth]{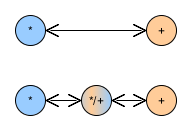}
    \hspace{0.15\textwidth}}
  \caption{Basic Ideas}
  \label{fig:basic-ideas}
\end{figure}

Figure~\ref{fig:basic-ideas} summarizes the basic ideas for this transformation:
To overcome the drastic fitness changes induced by changes in the tree
structure, the first simple ideas is to fix the tree structure
(Figure~\ref{fig:fixed-structure}) to a full (e.g binary) tree. This is
comparable to the limited tree depth or tree size that can often be found in
genetic programming. This limits the maximum change to the replacement of an
operator in the tree. However, directly switching from e.g. an addition to a
multiplication can still have quite a large impact on the behavior of the
formula. Therefore, the second idea is to make this transition smooth too, as
illustrated in Figure~\ref{fig:smooth-transition}. A very simple way to achieve
this smooth transition is to simply use a weighted average over all possible
operators as shown in Equation~\ref{eq:weighted-average}, where the
$\mathrm{op}_i(x, y)$ are the possible operators and $\mathrm{op}(x, y)$ is the
overall operation result. In the simplest case, when only two operators are
available, i.e. addition and multiplication, only a single factor needs to be
tuned, i.e.
$\mathrm{op(x, y)} := \alpha \cdot (x+y) + (1 - \alpha) \cdot (x \cdot y)$.
\begin{equation}
  \label{eq:weighted-average}
  \textrm{op}(x, y) := \sum_i^{2^d-1} \alpha_i \cdot \textrm{op}_i(x, y)
\end{equation}

So far, this yields a smooth optimization problem for the operator choices
where only real values have to be adjusted. However, this simply replaces every
operator with a weighted average of other operators and make the formula much
more complicated. Therefore, another simple addition is necessary: The overall
fitness function is augmented with a penalty for undecided operator
choices. In the simplest possible case where two possible operators are chosen an
inverted quadratic function that peaks at 0.5 and intersects with the abscissa
at zero and one can be used. When more than two operations are available a different penalty
function is required. In this case the least penalty should occur when exactly
one operator is chosen and it should increase progressively the more weight other
operators are receiving. Therefore, the simple formula shown in
Equation~\ref{eq:op-penalty} can been used to steer the optimization towards a
unique operator choice.
\begin{equation}
  \label{eq:op-penalty}
  p_{\mathrm{op}}(\alpha) := \left(\sum_i^{2^d-1} \alpha_i - \max_i \alpha_i \right)/(2^d-2)
\end{equation}


Now that the structure and the operators of the tree can be selected using only
real values, the last remaining aspect is how to make a smooth choices between
variables. This task is a litte more complicated as the variables should also
have a weight attached. However, another simple solution can be applied as shown
in Equation~\ref{eq:variable-selection} where the operation in the leaf node is
selected in addition to the variables, and $n$ is the number of input variables
in $X$. Please note that one additional weight $\beta_{i,n+1}$is included to
allow a constant to be selected which gives the ability to ``mute'' parts of the
tree by selecting for example a multiplication with one or an addition with zero
for some subtree.
\begin{equation}
  \label{eq:variable-selection}
  \mathrm{op}(X) := \sum_i^{2^{d-1}} \alpha_i \cdot
    \mathrm{op}(\beta_{i, 1}\cdot X_1, \ldots, \beta_{i, n} \cdot X_n, \beta_{i,n+1})
\end{equation}
This yields quite a large number of variable weights. As the number of leaf
nodes increases exponentially with the tree depth and is further multiplied with
the number of input variables. For example a smooth symbolic regression problem
with ten variables and a tree depth of five has only 31 operator weights but 176
variable weights.

\begin{figure}[htbp]
  \centering
  \subfloat[angular]{
    \label{fig:angular-selection}
    \includegraphics[width=0.2\textwidth]{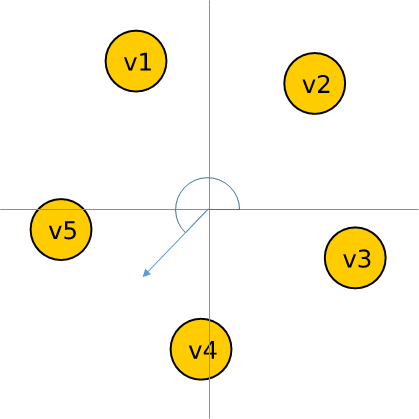}}
  \subfloat[nearest neighbor]{
    \label{fig:nearest-neighbor}
    \centering\includegraphics[width=0.3\textwidth]{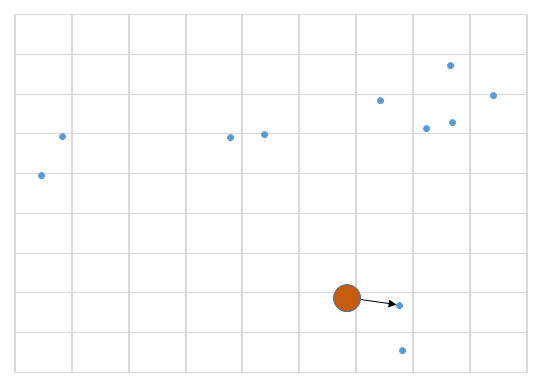}}
  \subfloat[weighted average]{
    \label{fig:weighted-average}
    \centering\includegraphics[width=0.3\textwidth]{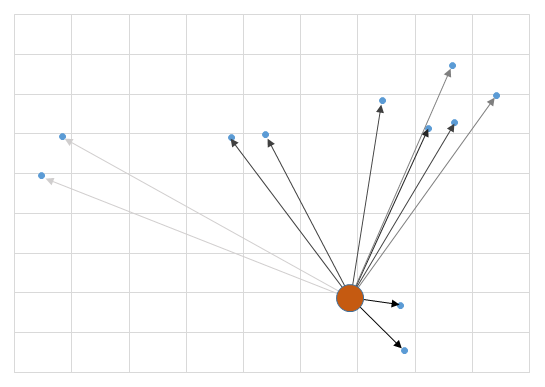}}
  \caption{Alternative Variable Selection Strategies}
  \label{fig:alt-var-sel}
\end{figure}

In Figure~\ref{fig:alt-var-sel} several alternative ideas are shown for variable
selection schemes with fewer resulting variables. However, they have not been
very promising in preliminary tests. Therefore, we can only recommend to use the
variable selection scheme with more variables, given the complexity of the
variable selection problem per se. The first idea was to use only a single angle
and take the two closest variables weighted by distance, as shown in
Figure~\ref{fig:angular-selection}, however, this blinds the algorithm by
completely hiding other choices. This might still be a good option for other
algorithms where diversity of the population is kept very high. Another idea was
to use multidimensional scaling \cite{Borg2005} to project the variables
according to their correlation onto e.g. a two dimensional plane and use only
two coordinates to choose between all
variables. Figure~\ref{fig:nearest-neighbor} shows the selection of the two
nearest neighbors, which has similar problems as the angular selection as it
completely hides other variable choices. The second alternative is to use again
a weighted average, however, using the distance to the coordinates as the
weights. 

It has to be noted, that also for the variable selection the optimization has to
be guided towards limiting the number of variables. This can be achieved
similarly to the operator choice, only that this time two or more non-zero
weights are acceptable and their weights can be subtracted from the penalty.

In summary, the new encoding transforms a problem with $2^d-1$ tree nodes, where
$d$ is the depth of the tree, $k$ possible operations at each node and $n$ input
variables into a problem with $(2^d-1)\cdot (k-1)$ operator weights $\alpha$ and
$(2^{d-1})\cdot n$ variable weights $\beta$. So, for example, for a depth of
five, with two operations and ten variables we get a 31+176=207 dimensional
real-valued problem instead of a combinatorial problem with $6\cdot 10^{20}$
possible choices. Obviously this does not make the problem less complex,
i.e. dimensions compared with choices, however it makes the fitness landscape much
smoother. One could think of discrete points in space in the combinatorial
formulation and filling the volume between them in the smooth approach.


\section{Experimental Results}
\begin{figure}[htbp]
  \centering
  \includegraphics[width=0.8\textwidth]{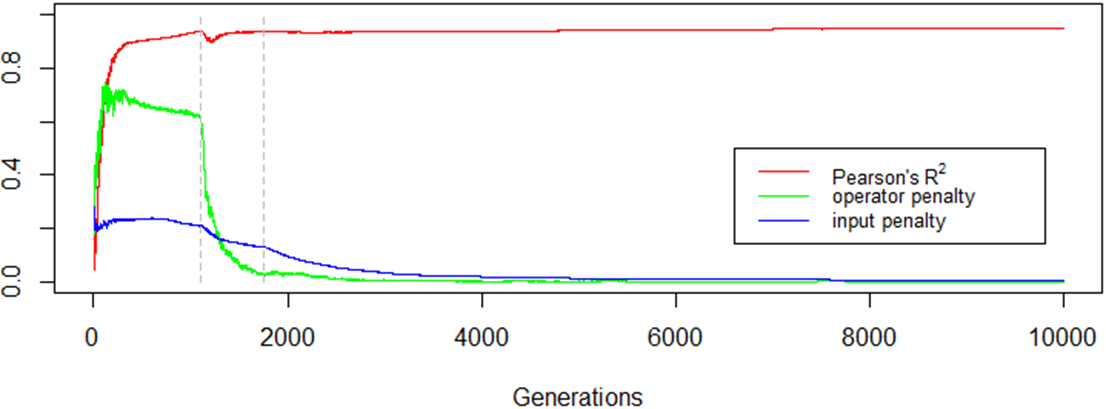}
  \caption{Results}
  \label{fig:results}
\end{figure}

We have tested this new implementation only on comparatively simple problems
most notably the \mbox{Poly-10}~Problem \cite{Langdon2008} using custom
operators in HeuristicLab \cite{Wagner2009} with a CMA Evolution
Strategy\cite{Hansen2006}. One very interesting aspect is shown in
Figure~\ref{fig:results} where the penalties for operators and variable weights
have been successively turned on. It can be seen that the correlation of the
formula slightly decreases as the optimization tries to lower the operator
penalty but quickly recovers with very low operator penalties, indicating a
crisp operator choice. This choice is also achieved rather quickly indicating
that it might not be so difficult to make this crisp operator choice. When
turning on the variable selection penalty a distinct knee and steeper slope can
be seen in the variable selection penalty curve, however, it is much harder to
decrease as many more weights are involved.

Finally, we have used the new formulation to calculate some fitness landscape
analysis measures. For the first time, traditional techniques can be applied to
symbolic regression problems and reasonable results can be obtained as shown in
Table~\ref{tab:fla}, where the fitness landscape of the \mbox{Poly-10}~Problem
was analyzed using different neighborhoods: In particular polynomial one
position or all position manipulators where used with contiguity of 15 or only 2
as well as a uniform one position manipulator. Both random and up-down walks
\cite{Jones1995} where performed to get a first impression of the landscape's
characteristics.

\begin{table}[htbp]
  \caption{Fitness Landscape Analysis of Symbolic Regression Problem}
  \label{tab:fla}
  \centering\scriptsize
\begin{tabular}{l|rrrrr}
                          & Poly-1-15 & Poly-All-15 & Poly-1-2 & Poly-All-2 & Uni-1   \\\hline
auto correlation          & 0.999     & 0.910       & 0.998    & 0.547      & 0.991   \\
corr. length              & 2245      & 57          & 1246     & 11         & 290     \\
density basin information & 0.628     & 0.619       & 0.626    & 0.593      & 0.628   \\
information content       & 0.546     & 0.394       & 0.686    & 0.403      & 0.399   \\
information stability     & 0.058     & 0.141       & 0.058    & 0.255      & 0.037   \\
partial inf. content      & 0.476     & 0.532       & 0.457    & 0.586      & 0.506   \\\hline
up walk length            & 328.063   & 124.872     & 14.321   & 5.179      & 83.433  \\
up walk len. variance     & 46668.196 & 5405.852    & 35.814   & 3.176      & 870.758 \\
down walk length          & 296.563   & 123.400     & 12.140   & 4.878      & 83.100  \\
down walk len. variance   & 27152.263 & 4627.477    & 26.232   & 2.713      & 832.024
\end{tabular}
\end{table}

\section{Conclusions}
While there is certainly still a lot of work needed to fine tune the optimization
process and play with different variants of variable selection and penalty
schemes, this transformation principle opens the door for classical fitness
landscape analysis applied to symbolic regression problems. The focus of future
work should therefore not be the tuning of algorithm performance but rather the
interpretation and utilization of FLA results generated for the class of
symbolic regression problems.

\subsubsection*{Acknowledgments}
The work described in this paper was done within the COMET Project Heuristic Optimization in Production and Logistics (HOPL), \#843532 funded by the Austrian Research Promotion Agency (FFG).

\bibliographystyle{splncs03}
\bibliography{SmoothGP}

\end{document}